\documentclass[letterpaper, 10 pt, conference]{ieeeconf}  
\IEEEoverridecommandlockouts                              
\usepackage{cite}
\usepackage{amsmath,amssymb,amsfonts}
\usepackage{algorithmic}
\usepackage{graphicx}
\usepackage{subfigure}
\usepackage{textcomp}
\usepackage{xcolor}
\usepackage{color}
\usepackage{siunitx}
\begin{document}

\title{\LARGE \bf Occlusion-Aware Path Planning for Collision Avoidance: Leveraging Potential Field Method with Responsibility-Sensitive Safety}

\author{Pengfei Lin$^{1}$, Ehsan Javanmardi$^{1}$, Jin Nakazato$^{1}$, and Manabu Tsukada$^{1}$
\thanks{$^{1}$P. Lin, E. Javanmardi, J. Nakazato, and M. Tsukada are with the Dept. of Creative Informatics, The University of Tokyo, Tokyo 113-8657, Japan. (e-mail: \{linpengfei0609, ejavanmardi, jin-nakazato, mtsukada\}@g.ecc.u-tokyo.ac.jp)}
}

\maketitle

\begin{abstract}
Collision avoidance (CA) has always been the foremost task for autonomous vehicles (AVs) under safety criteria. And path planning is directly responsible for generating a safe path to accomplish CA while satisfying other commands. Due to the real-time computation and simple structure, the potential field (PF) has emerged as one of the mainstream path-planning algorithms. However, the current PF is primarily simulated in ideal CA scenarios, assuming complete obstacle information while disregarding occlusion issues where obstacles can be partially or entirely hidden from the AV's sensors. During the occlusion period, the occluded obstacles do not possess a PF. Once the occlusion is over, these obstacles can generate an instantaneous virtual force that impacts the ego vehicle. Therefore, we propose an occlusion-aware path planning (OAPP) with the responsibility-sensitive safety (RSS)-based PF to tackle the occlusion problem for non-connected AVs. We first categorize the detected and occluded obstacles, and then we proceed to the RSS violation check. Finally, we can generate different virtual forces from the PF for occluded and non-occluded obstacles. We compare the proposed OAPP method with other PF-based path planning methods via MATLAB/Simulink. The simulation results indicate that the proposed method can eliminate instantaneous lateral oscillation or sway and produce a smoother path than conventional PF methods.
\end{abstract}


\section{Introduction}\label{intro}

Annually, approximately 1.3 million individuals tragically lose their lives due to road traffic crashes, while an additional 20 to 50 million people sustain non-fatal injuries, often resulting in long-term disabilities \cite{inju2021-rp}. Hence, autonomous vehicles (AVs) have been proposed to mitigate traffic injuries, leading to a safer driving environment. Collision avoidance (CA) is currently one of the most challenging tasks faced by AVs, specifically associated with the AV system's planning layer. Path planning in AVs involves the task of identifying an optimal and collision-free route that allows the vehicle to navigate through traffic while ensuring safety, comfort, and efficiency. Therefore, many excellent algorithms have been presented to complete the path-planning task \cite{Paden2016-ms, Claussmann2020-so}, including the rapidly-exploring random tree (RRT), A-star (A$^*$), dynamic window approach (DWA), potential field (PF), etc.

PF in path planning dates back several decades and has been widely studied in the field of robotics and other autonomous systems. The idea behind the PF is to model the environment as a field of attractive and repulsive forces, navigating the movement of a robot or AV. The basic principle of potential fields is that attractive forces pull the robot towards a goal location, while repulsive forces push it away from obstacles, and the resulting net force directs the robot or AV toward the goal while avoiding collisions \cite{Khatib1986-dv}. 

Although the PF method delivers the advantages of fast computation and a simple structure, its underlying premise assumes the availability of complete obstacle information. Thus, the performance of the PF may not meet expectations when handling the occlusion scenario where obstacles are partially or wholly invisible to the AV's sensors, which can provoke the inherent limitation of the PF (oscillations in the presence of obstacles) \cite{Koren1991-ar}. Therefore, this study introduces a novel occlusion-aware path planning (OAPP) method that combines responsibility-sensitive safety (RSS)-based PF to effectively address the challenges posed by the occlusion problem on expressway driving. The contributions of this study are briefly outlined below. 
\begin{itemize}
    \item We differentiate between obstacles perceived in visible and blind (occluded) areas and then utilize the RSS framework to conduct the violation check.
    \item We regulate the generation of the PF's virtual forces that are also used to produce the speed reference according to the violation check result. 
\end{itemize}
The remainder of this paper is structured as follows. Section II provides a review of related works on PF-based path-planning methods. Subsequently, Section III presents a detailed description of the proposed OAPP method. Section IV presents the comparative simulation results, followed by the conclusion in Section V.
\begin{figure*}[t]
    \centering
    \includegraphics[width=\hsize]{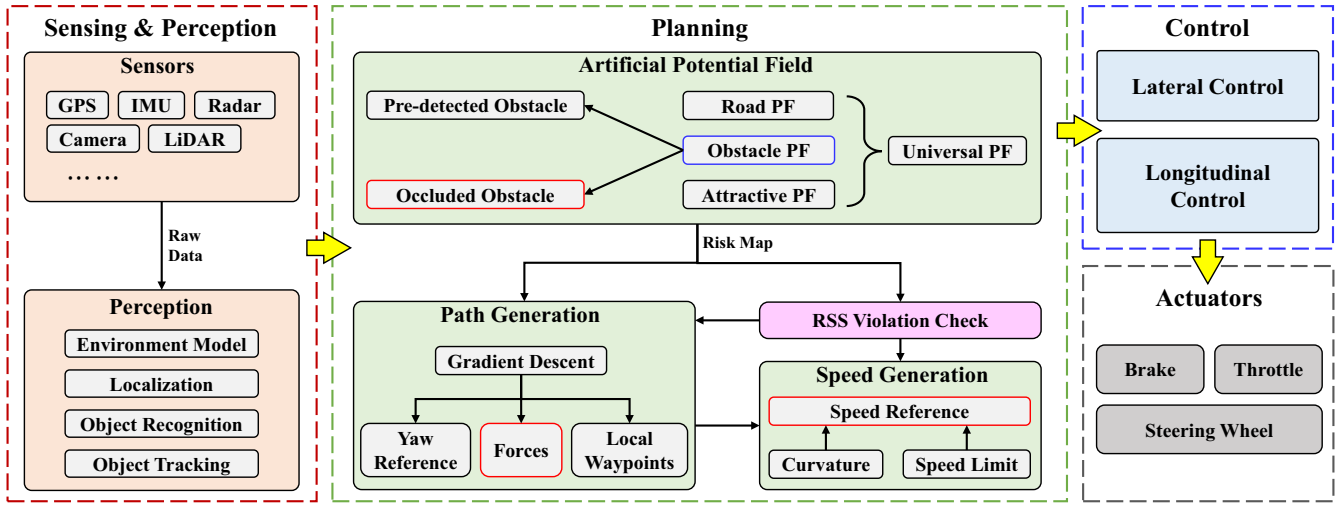}
    \caption{Proposed PF-OAPP framework consists of sensing \& perception, planning, control and actuators: Planning layer has the PF that can differentiate the visible and occluded obstacles, and then conduct the RSS violation check to adjust the virtual forces and generate the speed reference}
    \label{system_scheme}
\end{figure*}

\section{Related Work}\label{related_work}

In this section, we review the PF-based path-planning methods for AVs in the past decade. Date back to 2014, Shibata et al. \cite{Shibata2014-ib} proposed a velocity PF for the micro electric vehicle with the preview steering control method to deal with the CA problem. However, the simulation only considered a single obstacle driving slowly or being stationary. Then, Galceran et al. \cite{Galceran2015-oe} presented an integrated motion planning and control framework by using the PFs and torque-based steering actuation for a lower control effort. Similarly, the experimental stage showed only one obstacle in low-speed scenarios. Pongsathorn et al. \cite{Raksincharoensak2016-dn, Saito2016-qn} also presented a motion planning and control system focusing on PF-based risk optimization to avoid an occluded pedestrian that suddenly appeared behind a parked vehicle. Nevertheless, the study assumed that a virtual pedestrian was in the blind area and built the PF to achieve the pre-brake maneuver. Afterward, Ji et al. \cite{Ji2017-fk} and Rasekhipour \cite{Rasekhipour2017-yh} combined the PF with model predictive control (MPC) to provide more robust path planning and tracking control with the satisfaction of vehicle dynamics \cite{Rajamani2011-ua}. However, the obstacle information was assumed to be known in advance, and the PF's limitations were not discussed. Gao et al. \cite{Gao2019-mb} proposed multi-lane convoy control for AVs based on the distributed graph and the PF to resolve the instability of vehicle platoon. Still, the occlusion problem was not mentioned, and the vehicle dynamics were ignored. Later, Lin et al. \cite{Lin2020-ry, Lin2022-jm} applied the PF in waypoint tracking scenarios as a local path planning and used the clothoid curve to process the local waypoints to get a smooth path. However, the simulation also assumed the obstacle information was completely known, and only one obstacle was considered. Wu et al.~\cite{Wu2022-ae} presented a lane-change algorithm based on the PF that generates human-like trajectories, taking into account risks, drivers' focus shifts, and speed requirements, but their approach still required all obstacle information needs to be known in advance.

We also review recent work on planning methodologies for solving occlusion scenarios. Thornton et al. \cite{Thornton2018-eu} presented a modified value-sensitive design method for the AV's speed control to safely navigate an occluded pedestrian crosswalk. The research focused on a low-speed scenario around zebra crossings with a single longitudinal control strategy. {\c S}ahin et al. \cite{Tas2018-hv} proposed a specified motion planning given an uncertain environment model with occlusions. The simulation also concentrated on a T-junction with low-speed consumption and targeted only speed planning. Later, Naumann et al. \cite{Naumann2019-yu, Naumann2022-mn} provided safe but not overcautious and continuous-probabilistic motion planning strategies for intersections by considering the occlusion and limited sensor range. Still, the studies only sank into the low-speed scenario with speed generation while ignoring the path. Poncelet et al. \cite{Poncelet2020-me}, Wang et al. \cite{Wang2021-ks}, and Hang et al. \cite{Hang2022-ln} all dug into the unsignalized intersections with occlusion concerns and used different planning algorithms to resolve CA, including geometric method, hierarchical framework, and differential game approach. However, they all centered on an extremely low-speed driving scene where the vehicle drives under 10 m/s and discussed less on the path. Recently, Wang et al. \cite{Wang2023-yf} presented an occlusion-aware motion planning scheme that computed a static game tree considering the potential risk probability of the occluded area. But the research only validated the effectiveness of the proposed method within the vehicle kinematic under low-speed conditions. In the broader context of our study, our research aims to delve deeper into the implications of the occlusion problem on vehicle dynamics within high-speed driving scenarios, explicitly focusing on the utilization of the PF.

\section{Occlusion-Aware Path Planning}\label{pf}

This section introduces the principle of PF for collision avoidance and analyzes the problems encountered when applying the PF in the occlusion scenario with the proposed OAPP solution.

\subsection{Potential Field}

The PF encompasses various mathematical functions associated with the different elements of the road, such as road edges, lane dividers, and obstacles. These components contribute to the generation of virtual forces within the PF, including repulsive and attractive forces.

\subsubsection{Attractive Potential Function} In order to propel the ego vehicle forward, an attractive potential function $U_{AT}$ is designed to generate a tractive force.
\begin{equation}
    U_{AT}=
    \frac{1}{2}\lambda \left(X-X_{des}\right)^2,
    \label{attr_pf}
\end{equation}
where $\lambda$ represents the slope scale, while $X$ and $X_{des}$ refer to the longitudinal positions of the vehicle and the destination, respectively.

\subsubsection{Road Potential Function} The road potential function includes two parts: the road edges provide a clear boundary between the road and surrounding terrain, which helps prevent vehicles from veering off the road and potentially causing accidents. While lane dividers offer a physical barrier that helps prevent vehicles from drifting into adjacent lanes but can be easily violated if a lane-change decision is made. Therefore, we utilize the following formulas to describe the PFs of road edges $U_{RE}$ and lane dividers $U_{LD}$.
\begin{align}  
    U_{RE} &=
    \frac{1}{2}\xi(\frac{1}{Y-Y_{l,u}-\frac{l_w}{2}})^{2},\label{rel_re}\\
    U_{LD} &= 
    A_{LD}\exp{-\frac{(Y-Y_{LD}^i)^2}{2\sigma^2}},
    \label{rel_ld}
\end{align}

where $\xi$ represents the scaling coefficient, and $Y$ refers to the lateral position of the ego vehicle. The lateral positions of the lower and upper edges of the road are denoted by $Y_{l,u}$, respectively. $l_w$ represents the width of the vehicle, while $A_{LD}$ denotes the amplitude of the lane divider's potential function, and $Y_{LD}^i$ represents the lateral position of the $i^{th}$ lane divider. The variable $\sigma$ indicates the slope (rising or falling) of the lane potential.

\subsubsection{Obstacle Potential Function} The potential function for obstacles is designed to create a safety zone around them by assigning a high-risk value, thereby keeping the host vehicle at a safe distance from the obstacles. In 2017, Intel/Mobileye first proposed the RSS to define a set of mathematical rules that determine the safety boundaries for the AVs' movements \cite{Shalev-Shwartz2017-bh, Shashua2018-zn} that includes the safe longitudinal and lateral distances. And then, We can implement the RSS rules into the obstacle potential function $U_{OB}$, which is denoted as the following mathematical expression.
\begin{equation} 
    U_{OB}^j=
    \frac{\left|e^{-\sigma_{y}^j(X-X_{OB}^j)^{2}-\sigma_{x}^j(Y-Y_{OB}^j)^{2}}-\epsilon\right|}{\text{1}-\epsilon}
    \label{obstacle_pf}
\end{equation}
where
\begin{align*}
\centering
    \sigma_{x}^j=
    [ V\rho+&\frac{1}{2}a_{accel}^{max}\rho^2+\frac{(V+\rho a_{accel}^{max})^2}{2a_{brake}^{min}}-\frac{{V_{OB}^j}^2}{2a_{brake}^{max}} ]_+,\\
    \sigma_y^j=
    \zeta&+\Bigg[\frac{V^{lat}+V_{\rho}^{lat}}{2}\rho+\frac{{V_{\rho}^{lat}}^2}{2a_{brake}^{lat,min}}-\\
    &\left(\frac{V_{OB}^{lat}+V_{OB,\rho}^{lat}}{2}\rho+\frac{{V_{OB,\rho}^{lat}}^2}{2a_{brake}^{lat,min}}\right)\Bigg]_+,\\
    &V_{\rho}^{lat}=V^{lat}+\rho a_{accel}^{lat,max},\\
    &V_{obs,\rho}^{lat}=V_{obs}^{lat}+\rho a_{accel}^{lat,max}.
\label{rss}
\end{align*}
where $X$ represents the longitudinal position of the ego-vehicle, while $X_{OB}^j$ and $Y_{OB}^j$ denote the longitudinal and lateral positions, respectively, of the $j^{th}$ obstacle. $\sigma_{x}^i$ and $\sigma_{y}^i$ denote the weights assigned to the longitudinal and lateral distances between the ego vehicle and the obstacle, which are computed by the safe distances from the RSS rules. $\epsilon$ determines the threshold that defines the scope of the obstacle potential. $[\textit{h}]_+:\max\{\textit{h},0\}$. $V$ denotes the longitudinal speed of the ego-vehicle, while $\rho$ represents the response time required for detecting an emergency situation. $a_{accel}^{max}$ and $a_{brake}^{min}$ refer to the maximum acceleration and minimum braking capabilities of the ego-vehicle, respectively. The variable $V_{obs}^i$ denotes the longitudinal speed of the $i^{th}$ obstacle, and $a_{brake}^{max}$ represents the maximum braking capability of the obstacle. $lat$ represents the lateral motions of the vehicle, while $\zeta$ is the margin for fluctuations.
\begin{figure}[t]
    \centering
    \includegraphics[width=\hsize]{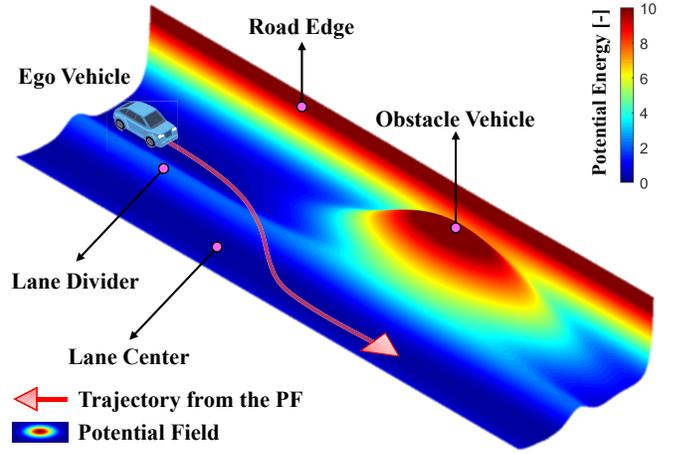}
    \caption{Universal PF of the road and the obstacle}
    \label{apf3d}
\end{figure}
\begin{figure*}[t]
    \centering
    \includegraphics[width=\hsize]{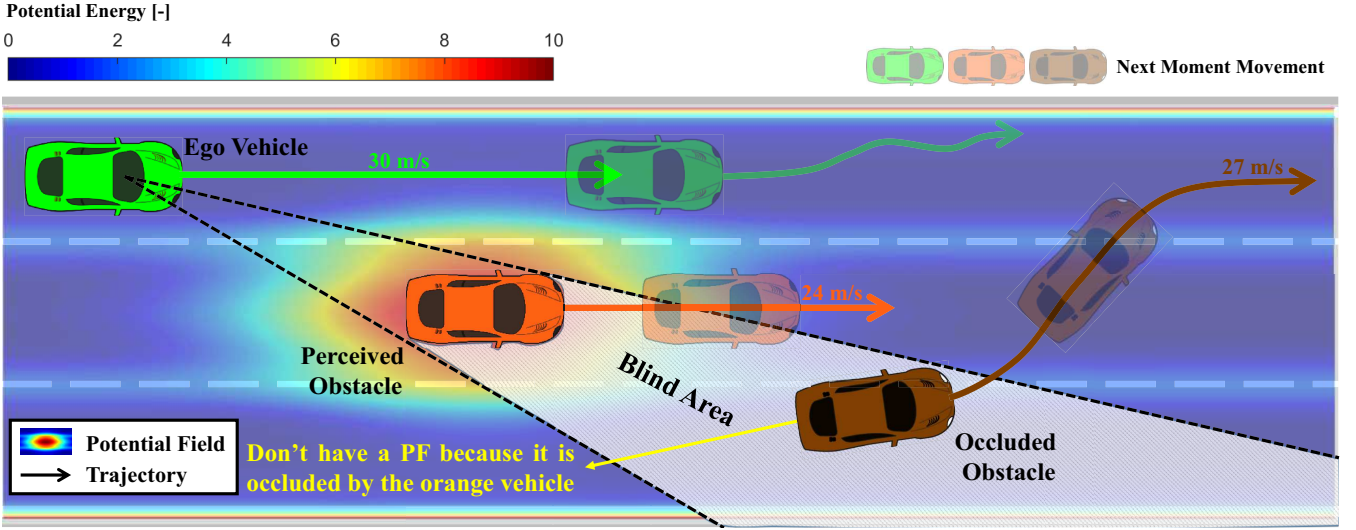}
    \caption{Occlusion problem: the brown vehicle is occluded by the orange vehicle from the view of the green vehicle, which the brown vehicle does not have a PF; the next moment movements of the AVs indicate the invisible driving maneuvers of the brown vehicle can impact the green vehicle's local path}
    \label{occlusion_explain}
\end{figure*}

\subsubsection{Universal Potential Field} By adding Eqs. (\ref{attr_pf})-(\ref{obstacle_pf}), we can obtain the universal PF $U_N$ (as depicted in Fig. \ref{apf3d}) and then apply the gradient descent method to get the virtual net force $\overrightarrow{F_{N}}$.
\begin{equation} 
    \overrightarrow{F_{N}}=-\!\nabla U_{N}=\overrightarrow{F_{AT}}+\overrightarrow{F_{RE}}+\overrightarrow{F_{LD}}+\overrightarrow{F_{OB}^j},
\end{equation}
where
\begin{equation*}
    U_N = U_{AT}+U_{RE}+U_{LD}+\sum_{j=1}^{j} U_{OB},
\end{equation*}
with $\overrightarrow{F_{AT}}$ is the virtual attractive force from Eq. (\ref{attr_pf}). $\overrightarrow{F_{RE}}$ and $\overrightarrow{F_{LD}}$ are the virtual repulsive forces from Eqs. (\ref{rel_re})-(\ref{rel_ld}). $\overrightarrow{F_{OB}^j}$ means the $j^{th}$ virtual repulsive force from Eq. (\ref{obstacle_pf}). Therefore, the ego vehicle can drive toward the target point while avoiding the obstacle under the interaction of these virtual forces, as depicted in Fig. \ref{force1}. Then, with the obtained net force $\overrightarrow{F_{N}}$, we can calculate the desired heading angle $\theta_{des}$ by the following equation.
\begin{equation}
    \theta_{des}=\arctan{\frac{\overrightarrow{F_{N}^{y}}}{\overrightarrow{F_{N}^{x}}}},
\end{equation}
with
\begin{equation*}
    \overrightarrow{F_{N}^{y}}=-\frac{\partial U_{N}}{\partial Y},\quad \overrightarrow{F_{N}^{x}}=-\frac{\partial U_{N}}{\partial X}.
\end{equation*}

\subsubsection{Occlusion Problem}

\begin{figure}
\centering
    \subfigure[Conventional PF method]{
        \begin{minipage}[t]{\linewidth}
            \centering
            \includegraphics[width=\hsize]{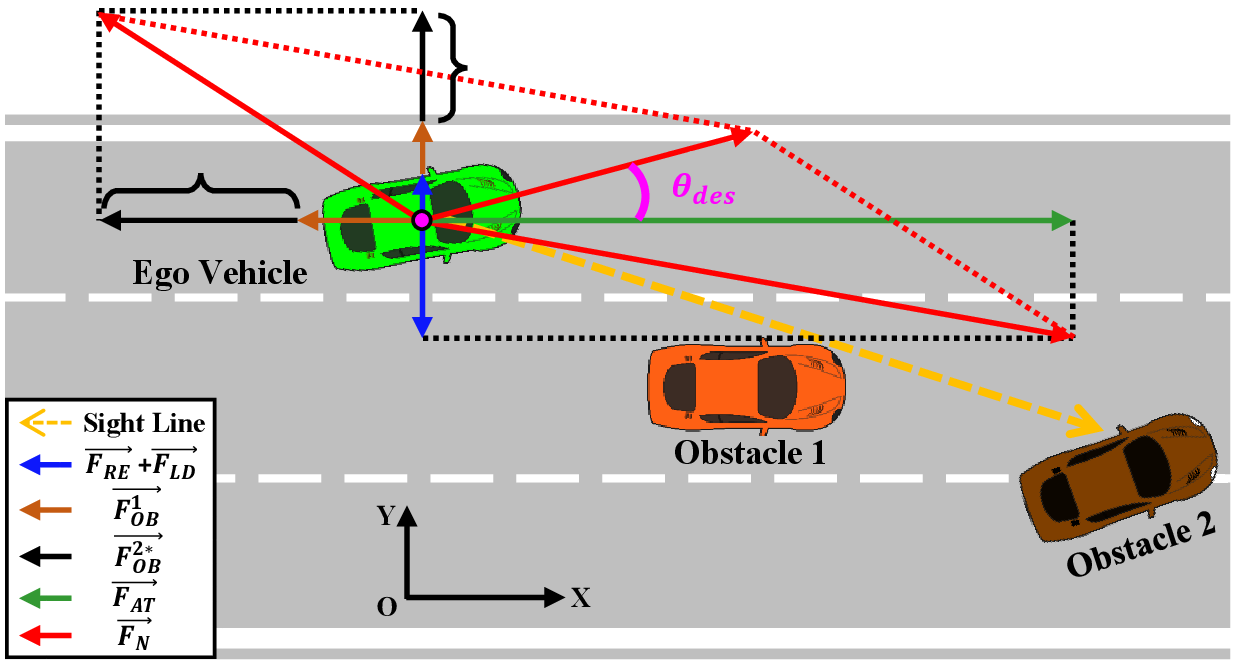}
        \end{minipage}
    \label{force1}
    }
    
    \subfigure[Proposed PF-OAPP method]{
        \begin{minipage}[t]{\linewidth}
            \centering
            \includegraphics[width=\hsize]{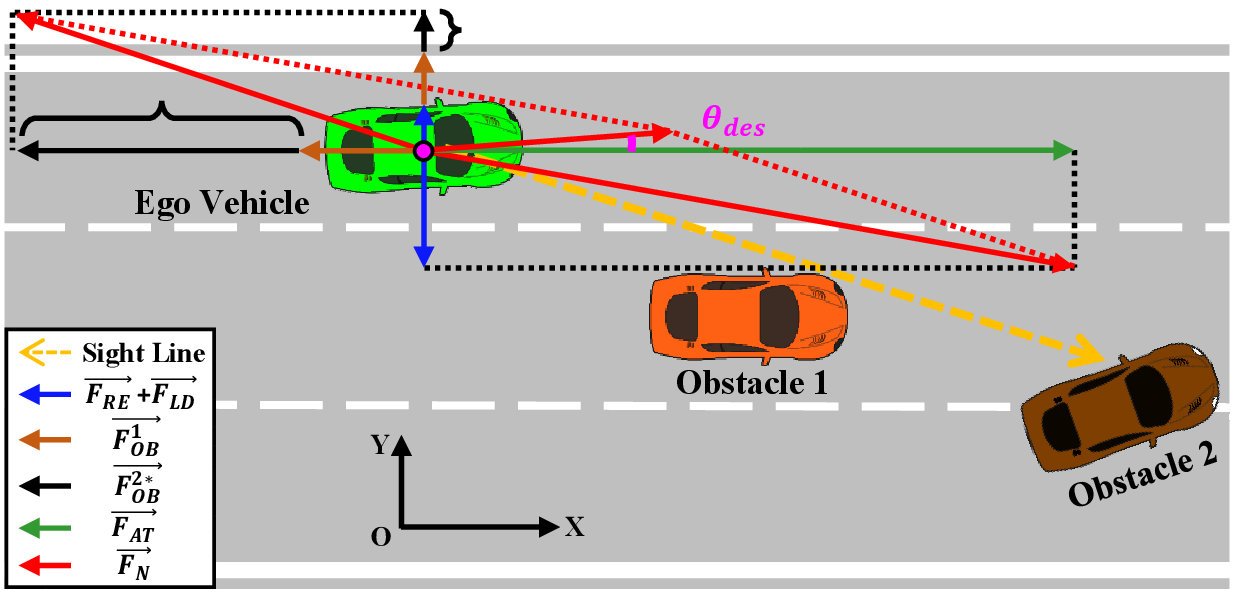}
        \end{minipage}
    \label{force2}
    }
\caption{Synthesis and decomposition of virtual forces when seeing the occluded obstacle 2 from two approaches}
\end{figure}

The scenario illustrated in Fig. \ref{occlusion_explain} portrays a situation on the expressway involving three vehicles operating independently at varying speeds. Within this visual representation, the ego vehicle (depicted in green) is observed to occupy the first lane. In contrast, the other two vehicles, driven by human drivers (shown in orange and brown), are positioned in the second and third lanes, respectively.It is important to highlight that the occlusion phenomenon occurs when the orange vehicle obstructs the brown vehicle, resulting in the brown vehicle becoming invisible from the viewpoint of the ego vehicle. This occlusion occurs within a designated blind area, distinctly marked as a shaded region in Fig. \ref{occlusion_explain}. Hence, the driving behaviors of the brown vehicle remain completely undetectable by the sensors of the ego vehicle, resulting in the absence of any established potential field (PF) between them. Suppose, at the next moment, the brown vehicle initiates an abrupt and continuous two-lane lane change maneuver. In such a scenario, the ego vehicle detects the presence of the brown vehicle and promptly establishes a potential field (PF) in response. However, as a consequence of the abrupt detection of the obstacle, the PF that arises between the ego vehicle and the brown vehicle can give rise to an excessive virtual repulsive force exerted on the ego vehicle. This force, depicted in Figure \ref{force1}, induces an immediate oscillatory impact on the ego vehicle, further amplifying the inherent instability in its motion.

To tackle the aforementioned problem, our proposed solution involves a series of steps. Firstly, we implement a labeling process to differentiate between obstacles originating from visible areas and those from blind areas. Specifically, for occluded obstacles, we utilize the principles of RSS to perform a violation check precisely at the moment they are perceived. In order to quantify the relative distance between the ego vehicle and the occluded vehicles, we introduce the variables $D_{rela}^x$ and $D_{rela}^y$, representing the longitudinal and lateral distances, respectively. Drawing from the guidelines outlined in the Saskatchewan Driver's Handbook \cite{saskatchewan-hb}, recommended maneuvers for emergency driving consist of a two-step approach. The initial step involves applying brakes to ensure longitudinal safety, followed by steering actions to avoid a collision on the expressway effectively. Therefore, to align our calculations with the RSS regulations and the suggested appropriate responses, we introduce the following adjustments to determine the virtual forces accurately. 
\begin{align}
    \overrightarrow{F_{N}^{x}}&=-(\frac{\partial U_{AT}}{\partial X}+\frac{\partial U_{RE}}{\partial X}+\frac{\partial U_{LD}}{\partial X}+\frac{\partial U_{OB}^j}{\partial X}+\alpha_1\frac{\partial U_{OB}^h}{\partial X}),\\
    \overrightarrow{F_{N}^{y}}&=-(\frac{\partial U_{AT}}{\partial Y}+\frac{\partial U_{RE}}{\partial Y}+\frac{\partial U_{LD}}{\partial Y}+\frac{\partial U_{OB}^j}{\partial Y}+\alpha_2\frac{\partial U_{OB}^h}{\partial Y}),
\end{align}
where
\begin{align*}
    \alpha_1&=
    \begin{cases}
        1-\frac{\pi}{2}\arctan{\frac{D_{rela}^x-\sigma_x^h}{\sigma_x^h}}& \text{if}\;D_{rela}^x\geq \sigma_x^h\\
        1+\frac{\pi}{2}\arctan{\frac{\sigma_x^h-D_{rela}^x}{\sigma_x^h}}& \text{if}\;D_{rela}^x< \sigma_x^h
    \end{cases}\\
    \alpha_2&=
    \begin{cases}
        1-\frac{\pi}{2}\arctan{\frac{D_{rela}^y-\sigma_y^h}{\sigma_y^h}}& \text{if}\;D_{rela}^y\geq \sigma_y^h\\
        \frac{\pi}{2}\arctan{\frac{\sigma_y^h-D_{rela}^y}{\sigma_y^h}}& \text{if}\;D_{rela}^y< \sigma_y^h
    \end{cases}
\end{align*}
\begin{figure*}[ht]
    \centering
    \includegraphics[width=\hsize]{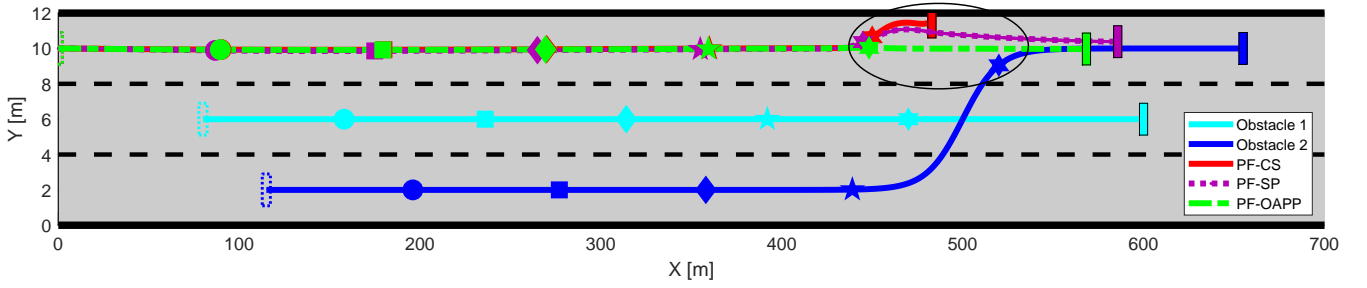}
    \caption{Trajectories of the CPF-based and OPF-based planners under the speed of 108 km/h}
    \label{path_2D}
\end{figure*}
\begin{figure*}
\centering
    \subfigure[Sideslip angle]{
        \begin{minipage}[t]{0.49\linewidth}
            \centering
            \includegraphics[width=\hsize]{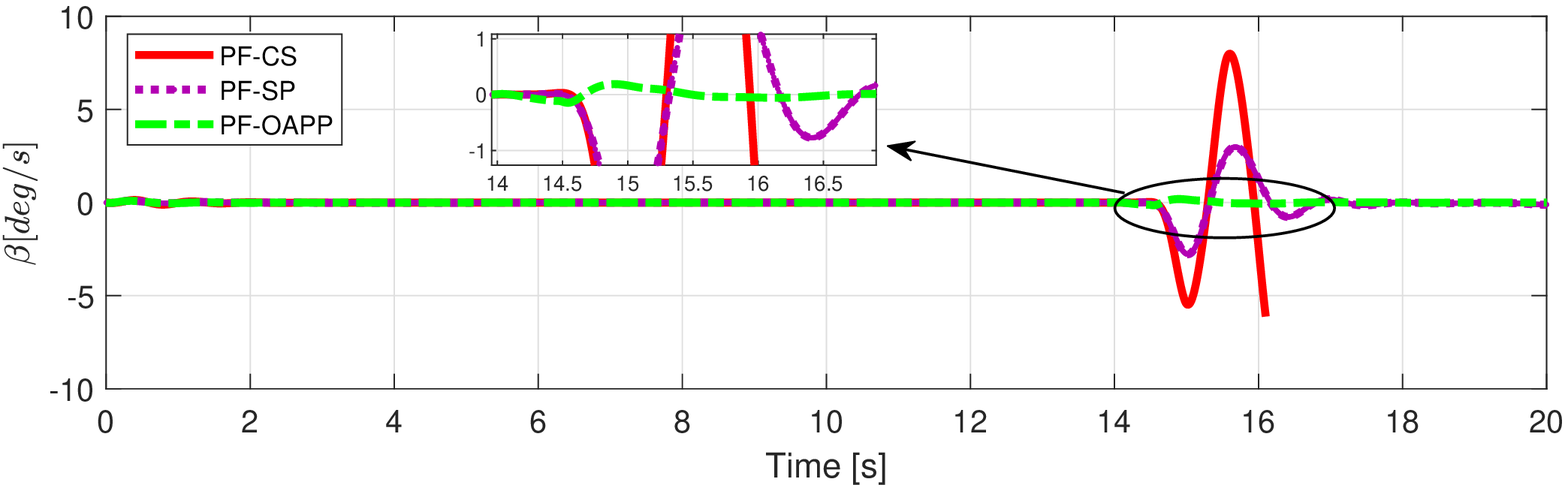}
        \end{minipage}
    \label{beta}
    }%
    \subfigure[Yaw angle]{
        \begin{minipage}[t]{0.49\linewidth}
            \centering
            \includegraphics[width=\hsize]{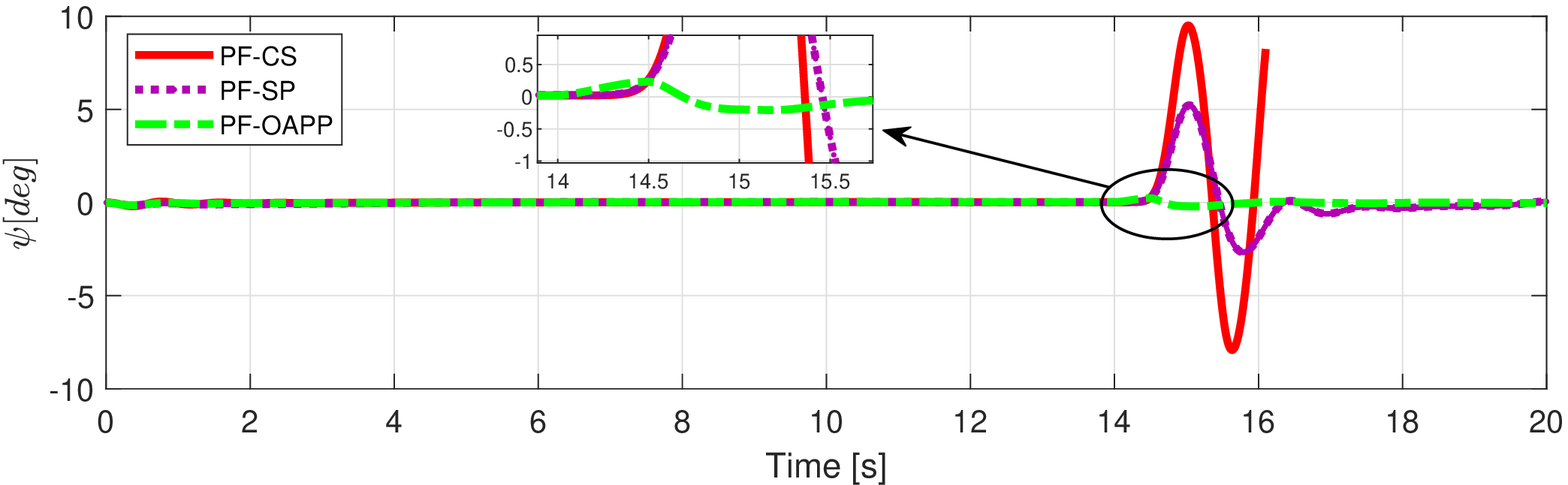}
        \end{minipage}
    \label{psi}
    }

    \subfigure[Front tire steering angle]{
        \begin{minipage}[t]{0.49\linewidth}
            \centering
            \includegraphics[width=\hsize]{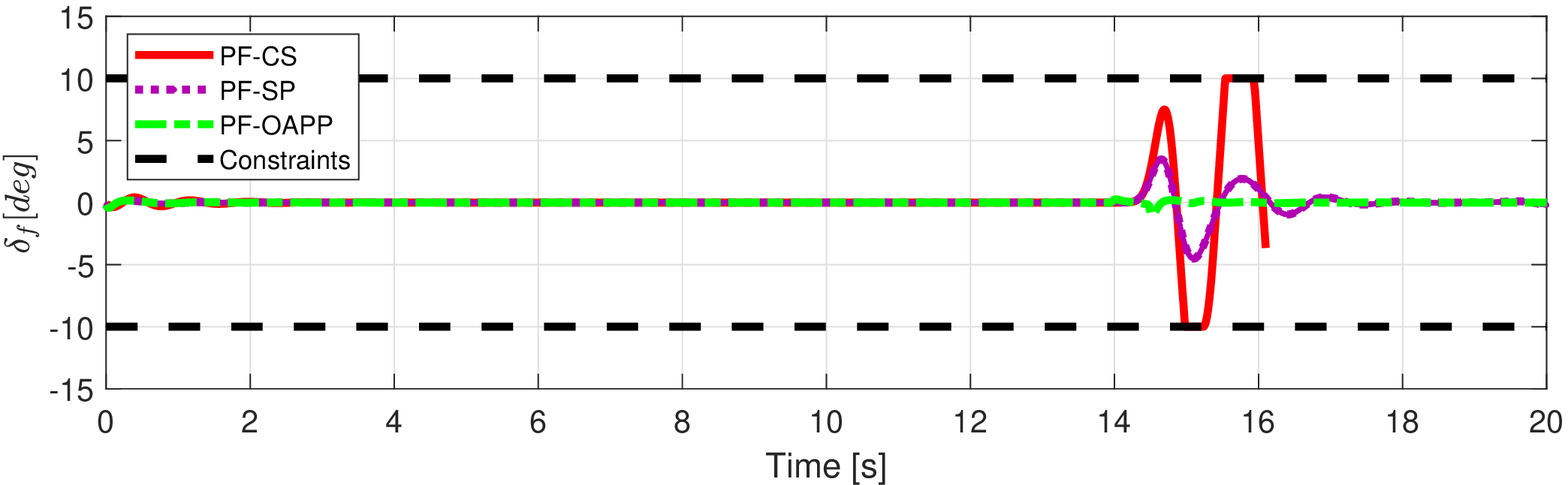}
        \end{minipage}
    \label{tire_steer}
    }%
    \subfigure[Longitudinal speed]{
        \begin{minipage}[t]{0.49\linewidth}
            \centering
            \includegraphics[width=\hsize]{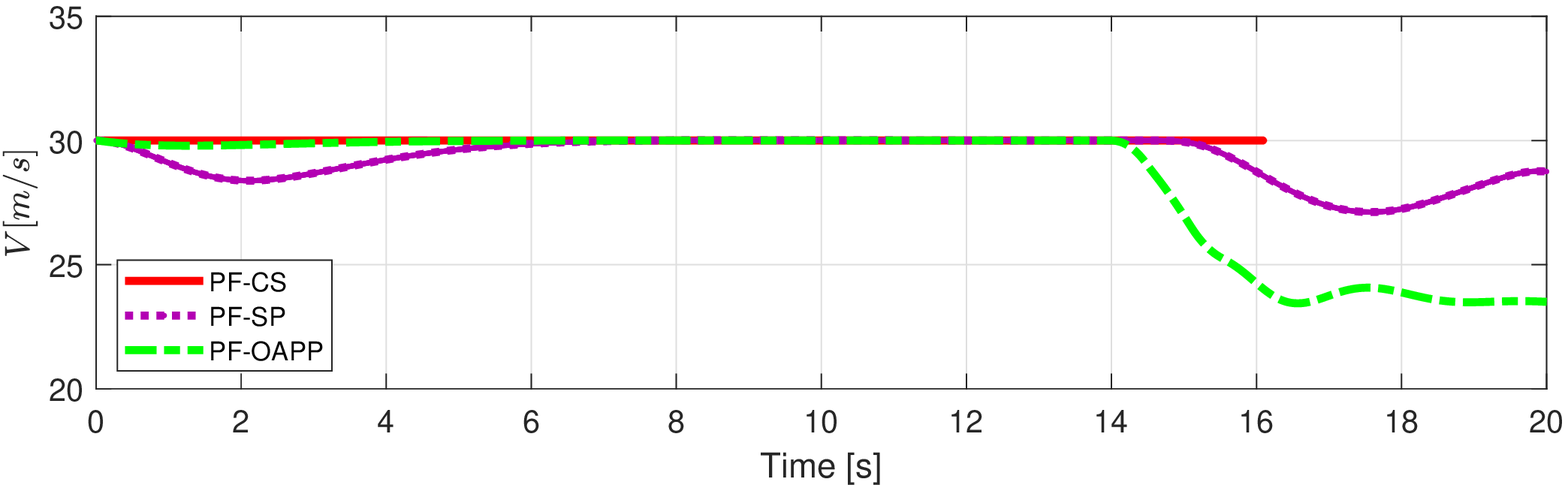}
        \end{minipage}
    \label{long_vel}
    }
\caption{Motion states of the obstacle vehicle based on PF-CS, PF-SP, and PF-OAPP, respectively}
\label{motion_states}
\end{figure*}
with $U_{OB}^h$ denotes the $h^{th}$ occluded obstacle that we mark from the blind area. $\alpha_{1,2}$ are the proportional coefficients based on the RSS safe distances with the normalization of the arctangent function. Therefore, we can mitigate the instant lateral oscillation or sway by using Eq. (7)-(8) to adjust the virtual forces while conforming to the safety criteria. Consequently, we can decrease the lateral virtual forces while increasing the longitudinal virtual forces when computing the PF for the occluded obstacle, as illustrated in Fig. \ref{force2}. Next, we use the adjusted forces to generate the longitudinal speed reference.
\begin{equation}
    \overrightarrow{F_{N}^x}-\overrightarrow{F_{C}}=M_va_{des},
\end{equation}
where $\overrightarrow{F_{C}}$ is the virtual criterion force that we can get when the ego vehicle drives at a constant speed without obstacles. $M_v$ is the virtual mass, and $a_{des}$ is the desired acceleration. Besides, we should also consider the speed limit on the expressway and the maximum speed for curved paths with the conformation of vehicle dynamics when calculating the reference speed $V^*$ for the controller.
\begin{equation}
    V^* = \min(V_{max},V_{des}),
\end{equation}
where
\begin{align*}
    V_{max} &= \min(V_{limit},\max(0,\sqrt{\frac{\mu g}{\kappa}})),\\
    V_{des} &= V_{cur}+a_{des}T,
\end{align*}
with $V_{limit}$ represents the limit speed on the expressway and $\mu$ is the frictional coefficient. $g$ means the gravitational acceleration and $V_{cur}$ denotes the current longitudinal speed of the ego vehicle. $T$ refers to the sampling time.

\section{Simulation Results}\label{AA}

\begin{figure}[t]
    \centering
    \includegraphics[width=\hsize]{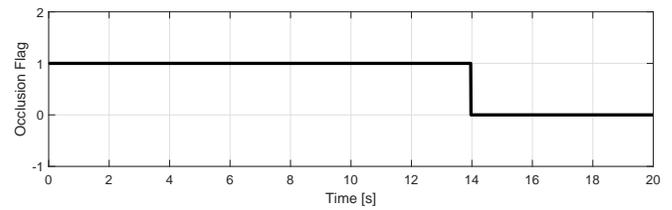}
    \caption{Occlusion flag: 1 means the occlusion appears; 0 means the occlusion disappears}
    \label{occlu_flag}
\end{figure}

In this section, we simulated an occlusion scenario on the expressway to inspect the proposed method via MATLAB/Simulink. Particularly, we have made an occlusion check that can output an occlusion flag, as shown in Fig. \ref{occlu_flag}. As described in Fig. \ref{occlusion_explain}, We first make two extended lines from the ego vehicle, and the extended lines will go through the edge corners of the orange vehicle. Then, we can use two linear equations to represent the two extended lines. Therefore, the occlusion check can be simplified to check the relative positions between the brown vehicle and two linear equations.

\subsection{Simulation Settings}

The basic settings of the occlusion scenario are also illustrated in Fig. \ref{occlusion_explain}. The ego vehicle is driving from an initial position of $(0,\;10)$ on the first lane, maintaining a speed of 30 m/s. Meanwhile, the orange vehicle is moving forward at a speed of 26 m/s, starting from an initial position on the middle lane at $(6,\;80)$. Additionally, the brown vehicle is shifting at a speed of 27 m/s on the third lane, located at position $(2,\;115)$. 

In the simulation study, we have conducted three comparative path planners to illustrate the performance of the proposed method: (\romannumeral1) PF-based path planner with constant speed (PF-CS) \cite{Lin2022-jm}; (\romannumeral2) PF-based path planner with speed planning (PF-SP) \cite{Wu2022-ae}; (\romannumeral3) Proposed PF-based occlusion-aware path planner (PF-OAPP). 

\subsection{Simulation Results and Analysis}

As shown in Fig. \ref{path_2D}, we can observe that both the paths of PF-CS (denoted as a red solid line) and PF-SP (denoted as a purple dotted line) have obvious lateral deviations, reaching maximum values of 11.46 m and 11.07 m, respectively. Besides, we discover that the PF-CS path planner stops the computation due to the trap to a local minimum. Conversely, the path of PF-OAPP (denoted as a green chain line) has no such lateral deviation, maintaining a stable driving status. But we can see that the path's length of PF-OAPP is shorter than that of PF-SP due to the greater deceleration, as depicted in Fig. \ref{long_vel}. The longitudinal speed of PF-CS decreases to 27.12 m/s while that of PF-OAPP reduces to 23.45 m/s. In addition, it should be noted that all the decelerated reactions appear after the brown vehicle is sensed at the time of $T=13.96$ s, where the occlusion flag changes from 1 to 0.

On the other hand, the motion states of the vehicles from different path planners are described in Fig. \ref{motion_states}. In Fig. \ref{beta}, we can find that the sideslip angle of PF-CS has a conspicuous oscillation from $T=14.62$ s with a maximum value of 7.99 $^\circ$/s. While the sideslip angle of PF-SP is smaller but still reaches a maximum value of 2.98 $^\circ$/s. On the contrary, the sideslip angle of PF-OAPP varies around zero, which indicates that there has been no significant sideslip occurring on the vehicle body. Similarly, we can observe the same phenomenon in the yaw angle that is depicted in Fig. \ref{psi}. The yaw angle of PF-CS exceeds 9$^\circ$ and has a short vibration from $T=14.28$ s to $T=16.11$ s while that of the PF-SP reaches 5.23$^\circ$ with less vibration. The yaw angle of PF-OAPP has a minimal change within 0.5$^\circ$, and no oscillations are found. Next, the front tire steering angle of the vehicle is presented in Fig. \ref{tire_steer} with a hard constraint of $\pm$10$^\circ$. We can see that the front tire steering angle of PF-CS reaches the constraint and oscillates between the hard bounds, while that of PF-SP has a negative maximum value of -4.51$^\circ$ and an attenuated oscillation from $T=14.28$ s to $T=17.50$ s. On the opposite, the front tire steering angle of PF-OAPP has a slight variation between 0.25$^\circ$ and -1.00$^\circ$.

\begin{figure}
\centering
    \subfigure[Longitudinal virtual force]{
        \begin{minipage}[t]{\linewidth}
            \centering
            \includegraphics[width=\hsize]{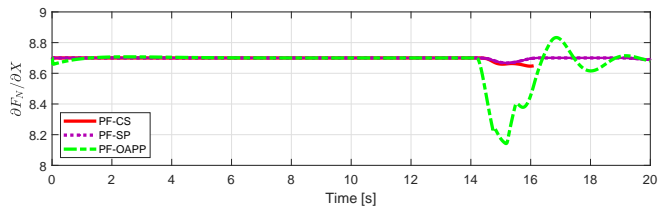}
        \end{minipage}
    \label{fx_t}
    }
    
    \subfigure[Lateral virtual force]{
        \begin{minipage}[t]{\linewidth}
            \centering
            \includegraphics[width=\hsize]{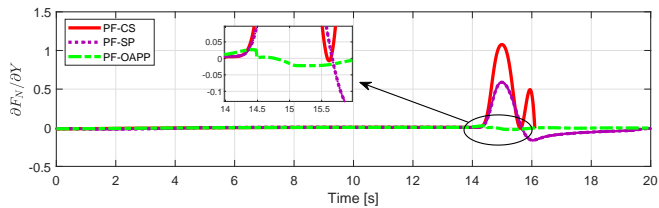}
        \end{minipage}
    \label{fy_t}
    }
\caption{Virtual forces from PF-CS, PF-SP, and PF-OAPP, respectively}
\label{forces}
\end{figure}

\begin{figure}[t]
    \centering
    \includegraphics[width=\hsize]{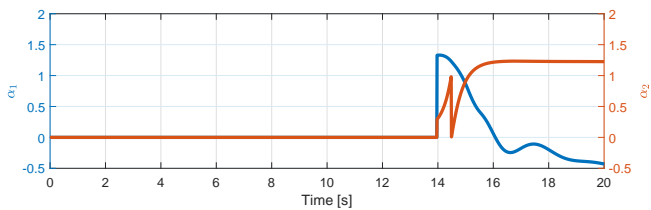}
    \caption{$\alpha_1$ and $\alpha_2$ from PF-OAPP}
    \label{alpha}
\end{figure}

Finally, the longitudinal and lateral virtual forces calculated from the three path planners are plotted in Fig. \ref{forces}. In Fig. \ref{fx_t}, we can see that the longitudinal virtual force of PF-OAPP starts to fall until reaching 8.14, while those of the other two planners have a slight change from 8.70 to 8.65 after the occlusion flag changes. From Fig. \ref{fy_t}, we can observe that PF-CS produces a more significant lateral virtual force, reaching a maximum of 1.08. At the same time, PF-SP also generates a lateral virtual force of 0.59. But the lateral virtual force of PF-OAPP varies around zero from 0.05 to -0.02. Furthermore, the proportional coefficients $\alpha_{1,2}$ of the PF-OAPP method are presented in Fig. \ref{alpha}. We can see that $\alpha_1$ exceeds 1 (reaches 1.33) and $\alpha_2$ is below 0.5 (around 0.4) when the occlusion flag changes from 1 to 0, which indicates that the PF-OAPP will produce a greater longitudinal virtual force and a smaller lateral virtual force than the current PF's values. 

\section{Conclusion}

In this paper, we have introduced a unique occlusion-aware path planning with the potential field, tackling the occlusion problem on the expressway. We have found that the occluded obstacle from the blind area will impose an unexpected virtual force from the PF modeling that can make the ego vehicle have a lateral sway. Therefore, we labeled the obstacle detected from the blind area and then used responsibility-sensitive safety to check the safety situation. After that, we can recompute the virtual forces based on the violation check. The simulation results have verified the effectiveness of the proposed PF-OAPP method compared to other PF methods in terms of eliminating lateral oscillation and generating a smoother path.

Our future research will prioritize investigating additional occlusion scenarios on the expressway, specifically caused by large trucks or tankers alongside or ahead of the ego vehicle due to their size and height. Firstly, we aim to further enhance the credibility and applicability of our proposed method by incorporating real vehicle experiments. Through these experiments, we will validate and refine our approach's effectiveness and reliability in real-world driving situations, gathering valuable data and insights to improve our research. We will also explore various occlusion scenarios and evaluate the impact of different environmental conditions, such as lighting and adverse weather, on our method's performance.

\section*{Acknowledgment}

These research results were obtained from the commissioned research Grant number 01101 by the National Institute of Information and Communications Technology (NICT), Japan, and the Japan Society for the Promotion of Science (JSPS) KAKENHI (grant number: 21H03423), and partly sponsored by the China Scholarship Council (CSC) program (No.202208050036) and JSPS DC program (grant number: 23KJ0391).
\bibliographystyle{IEEEtran}
\bibliography{Reference.bib}

\end{document}